\documentclass[letterpaper, 10 pt, conference]{ieeeconf}  

\IEEEoverridecommandlockouts                              

\usepackage[pdftex]{graphicx}
\graphicspath{{images/}{../jpeg/}{../pdf/}}
\DeclareGraphicsExtensions{.pdf,.jpeg,.png}

\usepackage{amsmath}

\usepackage{hyperref}

\title{\LARGE \bf
Evaluating ResNeXt Model Architecture for Image Classification
}

\author{Saifuddin Hitawala \\ David R. Cheriton School of Computer Science, University of Waterloo \\ saifuddin.hitawala@uwaterloo.ca 
}

\begin{document}

\maketitle
\thispagestyle{empty}
\pagestyle{empty}


\begin{abstract}

In recent years, deep learning methods have been successfully applied to image classification tasks. Many such deep neural networks exist today that can easily differentiate cats from dogs. One such model is the ResNeXt \cite{resnext} model that uses a homogeneous, multi-branch architecture for image classification. This paper aims at implementing and evaluating the ResNeXt model architecture on subsets of the CIFAR-10 \cite{cifar} dataset. It also tweaks the original ResNeXt hyper-parameters such as cardinality, depth and base-width and compares the performance of the modified model with the original. Analysis of the experiments performed in this paper show that a slight decrease in depth or base-width does not affect the performance of the model much leading to comparable results.

\end{abstract}

\section{Introduction}

The task of image classification as a problem gained importance in mid 2000s in the field of computer vision through the popularity of neural networks as well as through the availability of large image-based datasets. Many such natural and artificial image datasets and challenges exist today such as ImageNet \cite{imagenet}, CIFAR \cite{cifar}, STL \cite{stl}, SVHN \cite{svhn}, Pascal VOC \cite{pascalvoc} among others that help further advance the field of computer vision and neural networks. One of the most famous neural networks is the convolutional neural network that is used extensively for solving most of the visual recognition tasks if not all such as image classification, object detection, object localization, image segmentation, scene classification, face recognition and many more. Among these, one of the most well-researched problems is that of image classification.

With the availability of vast amount of labelled images and superior computing power, image classification has gained popularity within the computer vision community providing them a guideline for other tasks. This has led to deeper convolutional neural networks with varying architectures and model configurations. Usually, designing the architecture for these networks is quite difficult by hand as there are a lot of parameters and hyper-parameters involved. Thus, most of the popular neural network models such as the VGG-net \cite{vgg} are constructed by stacking multiple layers of similar blocks without any modifications to the block. Similarly, the optimal model configurations for such models are achieved either through trial and error or through a careful study and analysis of hyper-parameters and their effect on different parts of the network as specified in "Learning Deep Architectures for AI" \cite{learndeeparch}. One of the most recent and state-of-the-art models is the ResNeXt model proposed by Facebook AI Research in 2017.

The ResNeXt model is inspired from the VGG \cite{vgg} and ResNet \cite{resnet} models and has been desirable due to its improved performance for image classification tasks. This can be verified by the fact that the recent winner of the ImageNet Large Scale Visual Recognition (ILSVRC) 2017 Challenge \cite{imagenetchall} had their model's (SENet's) \cite{senet} Squeeze and Excitation module based on the ResNeXt model architecture. Thus, the popularity of ResNeXt model for image classification compels us to further investigate the model architecture and its configurations as well as improve the state-of-the-art in image classification.

In this paper, the originally proposed ResNeXt model is implemented using PyTorch \cite{pytorch}. Furthermore, some of the hyper-parameters of the model are tweaked and compared to the original model architecture. These tweaks are done by keeping all the hyper-parameters same except tweaking one of them to see its effect on the performance of the entire model. With the availability of many such hyper-parameters and a wide range of values, only some of these configurations have been evaluated and compared to the original model in this paper. 

The paper is organized in the following manner. Section II reviews work related to performing the task of image classification using deep neural networks. Section III gives a background on the ResNeXt model explaining its architecture, methodology and the hyper-parameters. Section IV gives on overview of the implementation details and the tweaks performed within the context of this paper. Section V provides an empirical evaluation of the results obtained after applying the tweaks to the original ResNeXt model on the CIFAR dataset. Finally, Section VI concludes this paper suggesting some of the directions for future work.

\section{Related Work}

There is a vast literature on image classification using neural networks. \textbf{LeNet} \cite{lenet} was one of the first convolutional neural networks that propelled the field of neural networks. The LeNet architecture was quite fundamental, in particular the insight that image features are distributed across the entire image, and convolutions with learnable parameters are an effective way to extract similar features at multiple location with few parameters. Moreover, the ability to save parameters and computation was a key advantage at the time since there was no GPU to help training, and even CPUs were slow.

After the popularity of deep neural networks, \textbf{AlexNet} \cite{alexnet} was released which was a deeper and much wider version of the LeNet and won by a large margin the difficult ImageNet competition. Some of the significant contributions of AlexNet were 1) the usage of rectified linear units (ReLU) \cite{relu} as non-linearities, 2) use of dropout technique to selectively ignore single neurons during training, a way to avoid overfitting of the model, 3) overlapping max pooling, avoiding the averaging effects of average pooling and 4) usage of NVIDIA GTX 580 GPUs to reduce training time. 

Then, in December 2013 the \textbf{Overfeat} \cite{overfeat} model (a derivative of AlexNet) was proposed by Yann LeCun from the NYU lab. The paper proposed learning bounding boxes, which later gave rise to many other papers on the same topic.

Soon after, \textbf{VGG} \cite{vgg} networks were proposed by Oxford and they were the  first to use much smaller 3x3 filters in each convolutional layers and also combined them as a sequence of convolutions. Instead of the 9x9 or 11x11 filters of AlexNet, filters started to become smaller, too dangerously close to the infamous 1x1 convolutions that LeNet wanted to avoid, at least on the first layers of the network. But the great advantage of VGG was the insight that multiple 3x3 convolutions in sequence can emulate the effect of larger receptive fields, for example 5x5 and 7x7. These ideas were later to be used in more recent network architectures such as Inception and ResNet.

\textbf{Network-in-network} \cite{nin} (NiN) was another model that followed similar principles as VGG net in terms of smaller filters. NiN had the great and simple insight of using 1x1 convolutions to provide more combinational power to the features of a convolutional layer. The NiN architecture used spatial MLP layers after each convolution, in order to better combine features before another layer. NiN also used an average pooling layer as part of the last classifier, another practice that would soon become common among CNN architectures.

In Fall 2014, \textbf{GoogLeNet} \cite{googlenet} was proposed by researchers at Google that aimed at reducing the computational burden of deep neural networks. The idea was to use an "Inception" module which at a first glance is basically the parallel combination of 1x1, 3x3, and 5x5 convolutional filters. But the great insight of the inception module was the use of 1x1 convolutional blocks to reduce the number of features before the expensive parallel blocks. This is commonly referred to as “bottleneck” today and used in future Inception models as well as the ResNeXt model.

\textbf{ResNets} \cite{resnet} were proposed in December 2015 and since then have created major improvements in accuracy in many computer vision tasks. The ResNet architecture was the first to pass human level performance on ImageNet, and their main contribution of residual learning is often used by default in many state-of-the-art networks today. Some of the key contributions of the ResNet model were as follows: \\
1) Showed that a naive stacking of layers to make the network very deep won’t always help and can actually make things worse. \\
2) To address the above issue, they introduced residual learning with skip-connections. The idea is that by using an additive skip connection as a shortcut, deep layers have direct access to features from previous layers. This allows feature information to more easily be propagated through the network. It also helps with training as the gradients can also more efficiently be back-propagated. \\
3) The first “ultra deep” network, where it is common to use over 100-200 layers.

Shortcut connections were then taken to the extreme with the introduction of \textbf{DenseNets} \cite{densenet}. DenseNets extended the idea of shortcut connections but had much more dense connectivity than ResNets. DenseNets connected each layer to every other layer in a feed-forward fashion. This allowed for each layer to use all of the feature-maps of all preceding layers as inputs, and its own feature-maps were used as inputs into all subsequent layers. DenseNets were shown to perform better than ResNets as they helped with alleviating the vanishing-gradient problem \cite{vanishgrad}, strengthening feature propagation, encouraging feature reuse, and substantially reduced the number of parameters.

Later on, \textbf{ResNeXt} model was proposed by Facebook that won 2nd place in ILSVRC 2016 classification task and also showed performance improvements in COCO detection \cite{coco} and ImageNet-5k set than their ResNet counter part. The ResNeXt model exposed a new dimension called cardinality as an essential network parameter, in addition to depth and width. Cardinality was demonstrated to be more effective than going deeper or wider when increasing model capacity, especially when increasing depth and width led to diminishing returns. Here, the ResNeXt model and its modifications are implemented and the results are evaluated for image classification on CIFAR dataset. 

\section{Background}

This section provides a brief overview of the ResNeXt model architecture, specifically the bottleneck block and aggregated transformations, and compares ResNeXt with other models.

\subsection{Model Architecture}

ResNeXt was proposed as a variant of ResNet with the building block as shown in Fig. 1.

\begin{figure}[h]
\centering
\includegraphics[scale=0.352]{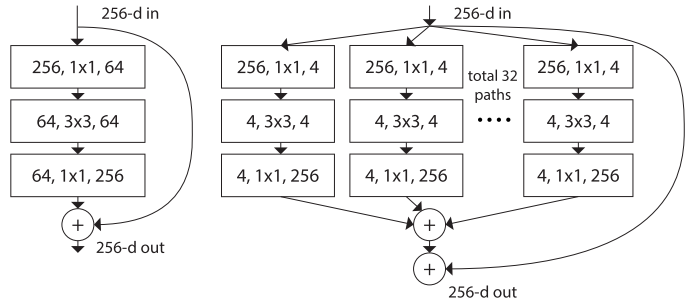}
\caption{Left: A block of ResNet. Right: A block of
ResNeXt with cardinality = 32, with roughly the same complexity.
A layer is shown as (\# in channels, filter size, \# out channels).}
\end{figure}

\begin{figure*}[h]
\centering
\includegraphics[scale=0.36]{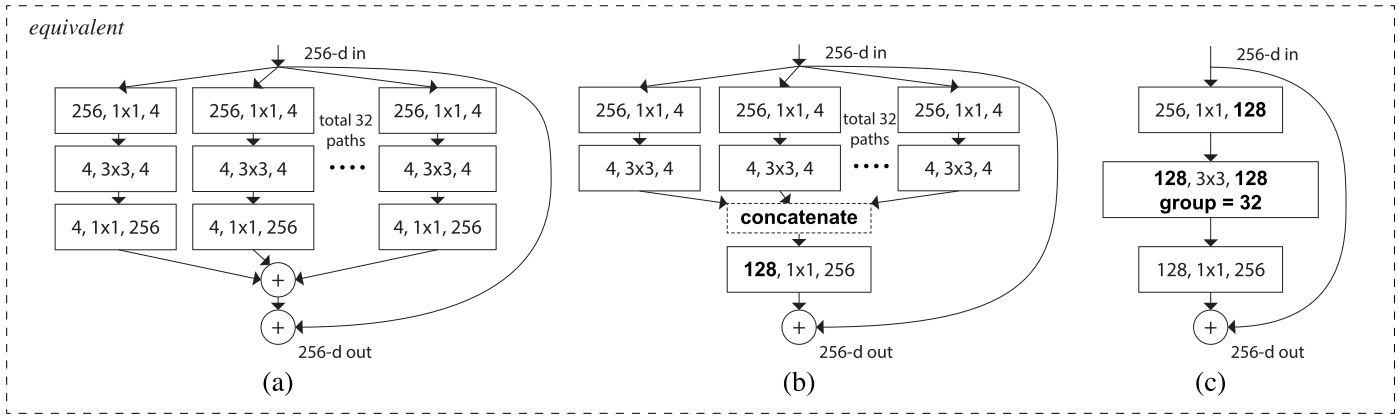}
\caption{Equivalent building blocks of ResNeXt. (a): Aggregated residual transformations, the same as Fig. 1(right). (b): A block equivalent
to (a), implemented as early concatenation. (c): A block equivalent to (a,b), implemented as grouped convolutions \cite{alexnet}. Notations in bold
text highlight the reformulation changes. A layer is denoted as (\# input channels, filter size, \# output channels).}
\end{figure*}

This block may look familiar as it is very similar to the Inception module \cite{inception}. They both follow the split-transform-merge paradigm, except in this variant, the outputs of different paths are merged by adding them together, while in Inception they are depth-concatenated.

The authors introduced a hyper-parameter called cardinality - the number of independent paths, to provide a new way of adjusting the model capacity. Experiments showed that accuracy can be gained more efficiently by increasing the cardinality than by going deeper or wider. 

This novel building block has three equivalent forms as shown in Fig. 2. In practice, the "split-transform-merge" is usually done by pointwise grouped convolutional layer, which divides its input into groups of feature maps and performs normal convolution respectively; their outputs are depth-concatenated and then fed to a 1x1 convolutional layer.

\subsection{Aggregated Transformations} 

The authors presented aggregated transformations as:

\begin{equation}
    \mathcal{F} (\mathrm{x}) = \sum_{i = 1}^{C} \mathcal{T}_i (\mathrm{x})
\end{equation}

where \(\mathcal{T}_i(\mathrm{x})\) is an arbitrary function. Analogous to a simple neuron, here \(\mathcal{T}_i\) projects \(\mathrm{x}\) into an (optionally low-dimensional) embedding and then transforms it.

In Eqn.(1), \(C\) is the size of the set of transformations to be aggregated. The authors refer to \(C\) as cardinality. The authors argue that the dimension of cardinality controls the number of more complex transformations.

The aggregated transformation in Eqn.(1) serves as the residual function (Fig. 1 right):

\begin{equation}
    \mathrm{y} = \mathrm{x} + \sum_{i = 1}^{C} \mathcal{T}_i (\mathrm{x})
\end{equation}

where \(\mathrm{y}\) is the output.

\subsection{Comparisons with Other Models}

\subsubsection{Inception-ResNet}

It can be seen that the module in Fig. 1(right) is equivalent to the module in Fig. 2(b). Thus, in Inception the outputs are depth concatenated while in ResNeXt the outputs from different paths are merged by adding them together. Another difference between the two models is that in Inception-ResNet each path is different (1x1, 3x3 and 5x5 convolution) from each other, while in ResNeXt, all paths share the same topology.

\subsubsection{Grouped Convolutions}

The usage of grouped convolutions dates back to the AlexNet paper, if not earlier. Grouped convolutions are supported by Caffe \cite{caffe}, Torch \cite{torch}, and other libraries, mainly for compatibility with AlexNet. A special case of grouped convolutions is the channel-wise convolutions in which the number
of groups is equal to the number of channels. The module in Fig. 2(a) becomes more succinct using the notation of grouped convolutions. This reformulation is illustrated in Fig. 2(c). Splitting can be essentially done by the grouped convolutional layer when it divides its input channels into groups. The block in Fig. 2(c) looks like the original bottleneck residual block in Fig. 1(left), except that Fig. 2(c) is a wider but
sparsely connected module.

\section{Technical Approach}

This section describes the dataset used and implementation details of the ResNeXt model. It also identifies the tweaks performed to the original ResNeXt model leading to various model configurations.

\subsection{Dataset}

The datasets used in the original ResNeXt paper were ImageNet-1k, ImageNet-5k, CIFAR-10, CIFAR-100 and COCO datasets. An attempt was done by the author of this paper at training the original ResNeXt model on CIFAR-10 dataset using SHARCNET's Graham cluster of 2 NVIDIA P100 Pascal GPUs provided by Prof. Ali Ghodsi. But, the training time was too long giving an estimated training time of 45 hrs for a complete training of the model. 

Due to the long training time for CIFAR-10 and limited timeline for the project, 3 subsets of the CIFAR-10 dataset were used. They are Cifar-2, Cifar-5 and Cifar-10 datasets (camel case denotes the dataset being a subset and number denotes the number of classes). Note that the CIFAR-10 dataset contains 50000 training images and 10000 test images of dimensions 32x32. Thus CIFAR-10 contains 5000 train images and 1000 test images of each category.

\subsubsection{Cifar-2}

The Cifar-2 dataset contains two categories - cats and dogs from the original CIFAR-10 dataset. The training set comprises of 2500 images of each category thus making the size of the entire training set equal to 5000 images. Similarly, the test set consists of 500 images of each category thus making it 1000 images in size.

\subsubsection{Cifar-5}

The Cifar-5 dataset contains five categories - cat, dog, deer, horse and frog from the original CIFAR-10 dataset. The training set comprises of 1000 images of each category thus making the size of the entire training set equal to 5000 images. Similarly, the test set consists of 200 images of each category thus making it 1000 images in size.

\subsubsection{Cifar-10}

The Cifar-10 dataset contains all the 10 categories from the original CIFAR-10 dataset. The training set comprises of 500 images of each category thus making the size of the entire training set equal to 5000 images. Similarly, the test set consists of 100 images of each category thus making it 1000 images in size.

\subsection{Implementation Details}

The implementation of ResNeXt model and its tweaks was done using PyTorch library and has been made publicly available on GitHub \footnote{The code can be found at \url{https://github.com/saifhitawala/ResNeXt}} online repository. The model is trained on the three datasets containing 5000 images mentioned above and validated on the 1000 images from the test set. The input image is 32x32 randomly cropped from a zero-padded 36x36 image or its horizontal flipping. Normalization is performed on the images and no other data augmentation is performed.

The first layer is a 3x3 conv. layer with 64 filters as in the original ResNeXt model. There are 3 stages each having 3 residual blocks as described in Section III. The network ends with a global average pooling and a fully-connected layer. The models are trained on SHARCNET's Graham cluster of 2 NVIDIA P100 GPUs with a mini batch size of 128, with a weight decay of 0.0005 and a momentum of 0.9. The training is started with a learning rate of 0.1 and the models are trained for 300 epochs, reducing the learning rate at the 150-th and 225-th epoch. All of these implementation details are similar to that used in the original ResNeXt paper.

\subsection{Tweaks Performed}

The ResNeXt model contains a lot of parameters and hyper-parameters that could be tweaked or modified. Some of the important hyper-parameters listed in the ResNeXt paper were cardinality (group size), model depth and base-width (channel capacity). These hyper-parameters have been experimented with keeping other hyper-parameters the same as can be seen as follows: 

\subsubsection{Cardinality}

The original ResNeXt model was evaluated on CIFAR datasets with cardinalities 8 and 16. Here, the cardinalities have been reduced to 1, 2 and 4 keeping other hyper-parameters the same and their models' performance has been compared to those trained with cardinalities 8 and 16 on Cifar-2, -5 and -10 datasets. These results can be seen in Section V.

\subsubsection{Depth}

For CIFAR datasets, the original ResNeXt model had a depth of 29 while it had a depth of 50 and 101 for the ImageNet datasets. Surely, increasing the number of layers usually improves the accuracy, but also increases the number of parameters as well as training time. As time was of a concern, experiments were performed reducing depth to 20 (number chosen as depth-2 must be divisible by 9 as per the bottleneck architecture) to evaluate if the accuracy degraded too much compared to the originally proposed depth. The results were not that different with a significantly reduced training time. All other hyper-parameters were kept the same. (cardinality=8, base-width=64)

\subsubsection{Base Width}

The original model was evaluated with base width of 64. Here, the base width was reduced to 32 keeping all other hyper-parameters same and evaluated on Cifar-2, -5 and -10 datasets. Reducing the width reduced training time considerably and the results can be seen in the following section.

\section{Experimental Results}

This section showcases the results obtained from evaluating the ResNeXt model and it's tweaks on the Cifar datasets.

\subsection{Varying Cardinality}

\begin{figure}[h]
\includegraphics[scale=0.314]{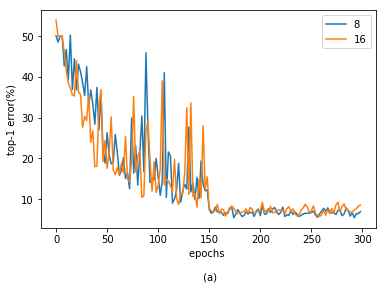}
\includegraphics[scale=0.314]{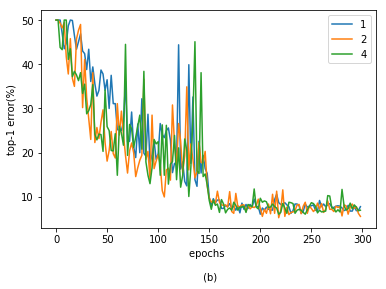}
\includegraphics[scale=0.314]{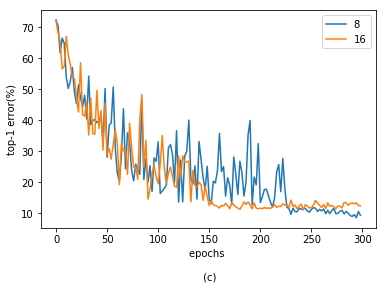}
\includegraphics[scale=0.314]{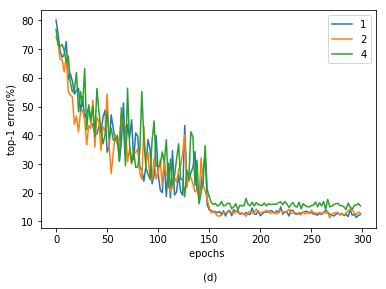}
\includegraphics[scale=0.314]{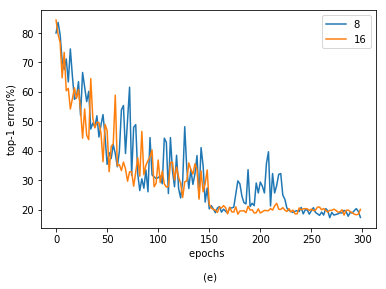}
\includegraphics[scale=0.314]{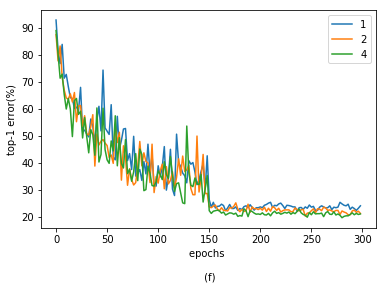}
\caption{\textbf{Cifar experiments on cardinality}. (a) and (b) show the test error against epochs for Cifar-2. Similarly, (c) and (d) show test error against epochs for Cifar-5 and (e) and (f) for Cifar-10.}
\end{figure}

Here, the cardinality of the bottleneck layers was varied from the original cardinalities of 8 and 16 to 1, 2 and 4. These models were evaluated on all the three datasets Cifar-2, -5 and -10 as can be seen Fig. 3.

Fig. 3 shows how test error varies with varying cardinalities keeping the model depth and base-width constant (29 and 64 respectively). Here, the graphs on the left (a), (c) and (e) were models trained with cardinality 8 and 16 whereas those on the right (b), (d) and (f) were trained with cardinality 1, 2 and 4. 

It can be inferred that for smaller cardinalities the training is initially slow and the model converges slowly as compared to those with higher cardinalities. Moreover, as the number of categories increases, the curves become smoother implying that the ResNeXt model is quite robust and is invariant to changes in data (Cifar-10 introduces categories with no relation to each other such as airplane and truck while Cifar-5 contains animal categories). Also, the models with cardinalities 1, 2 and 4 stop learning anything useful after the 160-th epoch and eventually have lower best test accuracies as compared to that with cardinality 8 and 16 as can be seen in Fig. 4.

\begin{figure}[h]
\includegraphics[scale=0.29]{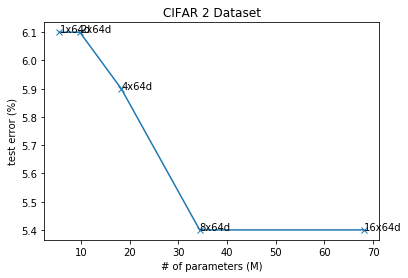}
\includegraphics[scale=0.29]{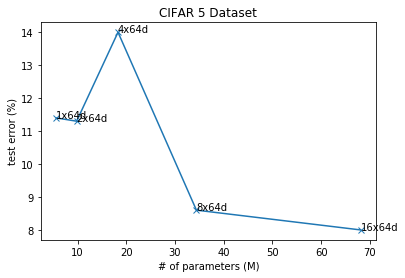} 
\centering
\includegraphics[scale=0.29]{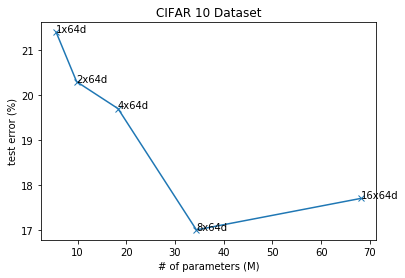}
\caption{Test error \textit{vs.} model size for Cifar-2, -5 and -10}
\end{figure}

Fig. 4 displays that as the cardinality increases the model size increases and usually the test error decreases. Although there are some exceptions to this (Cifar-5 cardinality 4 and Cifar-10 cardinality 16), it can be attributed to improper training as the models were trained in batches on SHARCNET due to limited job allocation time (max. 3 hours). Thus, each model was trained in 2-3 jobs and it is highly likely that one of these models was trained more than the required number of epochs due to improper load/resume from checkpoint.

Also, the best accuracy that we get for Cifar-10 dataset \footnote{Note: Not CIFAR-10 dataset. Remember, Cifar-10 is a subset of CIFAR-10.} is 83\% which is pretty close to that for CIFAR-100 (83.7\%) as mentioned in the original ResNeXt paper. This is due to the fact that both Cifar-10 and CIFAR-100 have 500 train images and 100 test images for each category thus leading to similar levels of accuracies. This implies that training size matters a lot.

\subsection{Varying Depth}

Here, the depth of the model was reduced from 29 to 20 thus reducing the number of bottleneck stages from 3 to 2. The performance of these models can be seen in Fig. 5.

\begin{figure}[h]
\includegraphics[scale=0.313]{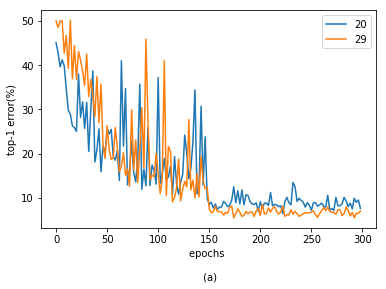}
\includegraphics[scale=0.313]{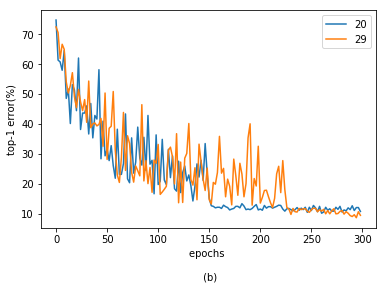} 
\centering
\includegraphics[scale=0.313]{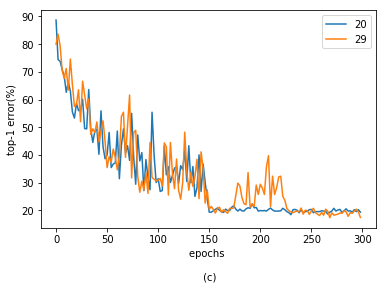}
\caption{\textbf{Cifar experiments on depth}. (a), (b) and (c) show results for Cifar-2, -5 and -10 respectively.}
\end{figure}

As it can be seen from Fig. 5, the final model accuracy for depth 20 is less than that with depth 29. This supports the idea that with an increase in depth the accuracy increases. However, this decrease in the final accuracy is not much compared to the significant reduction in training time as can be seen in Table 1. (Note that subset [1/10-th] of dataset has been used and the training time on entire dataset would be much large). Moreover, the variance in the test error for model with depth 20 is much less than the variance for model with depth 29.

\begin{table}[ht]
\caption{Table showing the Training Times for Different Models}
\label{table_example}
\begin{center}
\begin{tabular}{|c|c|c|}
\hline
Cardinalities & ResNeXt-29 (300 epochs) & ResNeXt-20 (300 epochs)\\
\hline
1x64d & \(\sim\)8.5 s/epoch (\(\sim\)45min) & \\
\hline
2x64d & \(\sim\)10 s/epoch (\(\sim\)50min) & \\
\hline
4x64d & \(\sim\)15 s/epoch (\(\sim\)80min) & \\
\hline
8x64d & \(\sim\)20 s/epoch (\(\sim\)105min) & \(\sim\)13 s/epoch (\(\sim\)70min) \\
\hline
16x64d & \(\sim\)40 s/epoch (\(\sim\)200 min) & \\
\hline
8x32d & \(\sim\)13.5 s/epoch (\(\sim\)75 min) & \\
\hline
\end{tabular}
\end{center}
\end{table}

\subsection{Varying Width}

Here, the base-width of the model (number of channels) was reduced from 64 to 32 reducing the number of input and output channels for each layer in the bottleneck block. This reduces the model complexity significantly from 32.4 million parameters to 22.8 million parameters. The performance of these models can be seen in Fig. 6.

\begin{figure}[h]
\includegraphics[scale=0.313]{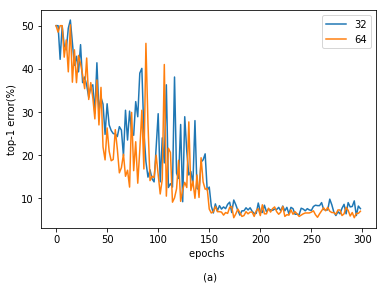}
\includegraphics[scale=0.313]{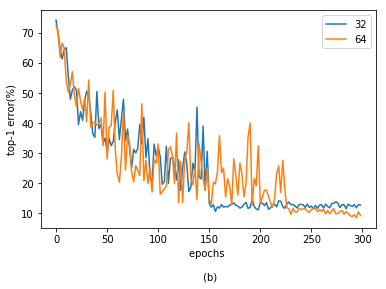} 
\centering
\includegraphics[scale=0.313]{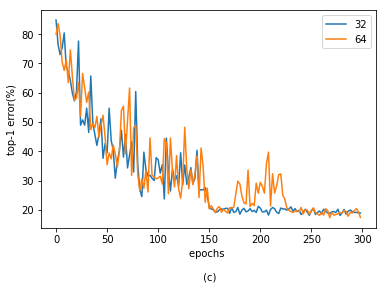}
\caption{\textbf{Cifar experiments on base-width}. (a), (b) and (c) show results for Cifar-2, -5 and -10 respectively.}
\end{figure}

Fig. 6 shows that models with base-width 32 perform comparably to models with base-width 64. Moreover, the variance in test accuracy as well as the convergence rate are also almost the same. Also, it can be seen from Table 1 that the total training time for model with base width 32 was \(
\sim\)75 mins which is \(\sim\)30 mins less than that with base width 64. This implies that training the model with base width 32 might in fact be a better idea if smaller sized models are to be trained in lesser time.

\section{Conclusion and Future Work}

The results in Sections V indicate that most of the hyper-parameters used in the original ResNeXt model configuration perform best. However, decreasing the depth and base-width might be a good idea if one needs to train a model of smaller size and in a shorter time period with comparable accuracy as that of the original ResNeXt model. Moreover, if accuracy is of utmost concern, then increasing the depth or base-width could be a good idea too.

As mentioned earlier, the ResNeXt model contains many parameters and  hyper-parameters with a wide range of values. Thus, this work could be researched further by experimenting with other parameters such as the learning rate, momentum, weight decay, etc. Also, the hyper-parameters such as depth and base-width could be increased in order to improve the model accuracy. Since, the models in this paper were evaluated on subsets of CIFAR-10 dataset, one could also evaluate them on complete datasets as well as on ImageNet and COCO datasets. Finally, the architecture of the model can be experimented with and combined with the architectures of other state-of-the-art models to construct a hybrid model architecture that performs well than the original models.

\addtolength{\textheight}{-12cm}   


\section*{Acknowledgement}

The author would like to thank Prof. Jeff Orchard for providing useful insights and directions to investigate while accepting the project proposal. His comments and feedback were helpful in modeling and implementing the project. The author would also like to thank Prof. Ali Ghodsi for providing access to SHARCNET clusters thus allowing him to perform all the experiments and reduce training time considerably.

\end{document}